\documentclass[journal]{IEEEtran}

\ifCLASSINFOpdf
\else
\fi
\usepackage[utf8]{inputenc} 
\usepackage[T1]{fontenc}    
\usepackage{hyperref}       
\usepackage{url}            
\usepackage{booktabs}       
\usepackage{amsfonts}       
\usepackage{nicefrac}       
\usepackage{microtype}      

\usepackage{color}
\usepackage{amsmath}
\usepackage{multirow}
\usepackage{graphicx}
\usepackage{pdflscape}

\usepackage{wrapfig}
\usepackage{lipsum}  

\usepackage{algorithm}
\usepackage{algorithmic}

\usepackage{times}
\usepackage{subfigure}

\usepackage{bbm}

\usepackage{amsthm}
\usepackage{amssymb}

\newtheorem{theorem}{Theorem}

\usepackage{epstopdf}
\usepackage{enumerate}

\newcommand{\tree}{{\mathcal T}}

\newcommand{\algs}{{\mathcal A}}

\newcommand{\alg}{{\mathcal A}}

\newcommand{\data}{{D}}

\newcommand{\dist}{{\mathcal D}}

\newcommand{\R}{{\mathbb R}}

\usepackage{array}
\newcolumntype{L}{>{\arraybackslash}m{6.2cm}}
\newcolumntype{M}{>{\centering\arraybackslash}m{1.8cm}}
\newcolumntype{N}{>{\arraybackslash}m{16cm}}

\begin{document}
    %
    \title{ToPs: Ensemble Learning with Trees of Predictors}
    %
    %
    %
    
    \author{Jinsung~Yoon, William~R.~Zame, and~Mihaela~van der Schaar,~\IEEEmembership{Fellow,~IEEE}
        \thanks{J. Yoon is with the Department of Electrical Engineering, University of California, Los Angeles, CA, 90095 USA e-mail: jsyoon0823@ucla.edu.}
        \thanks{W. R. Zame is with the Departments of Mathematics and Economics, University of California, Los Angeles, CA, 90095 USA e-mail: zame@econ.ucla.edu.}
        \thanks{M. van der Schaar is with the Department of Engineering Science, University of Oxford, OX1 3PJ UK e-mail: mihaela.vanderschaar@oxford-man.ox.ac.uk.}
        \thanks{This paper has supplementary downloadable material available at \url{http://ieeexplore.ieee.org}, provided by the author. The material includes additional experiment results, other related works and the details of experimental settings. This material is 1.0MB in size.}
    }

    \maketitle
    
    \begin{abstract} 
    We present a new approach to ensemble learning.  Our approach differs from previous approaches in that it constructs and applies different predictive models to different subsets of the feature space.  It does this by constructing  a tree of subsets of the feature space and associating a predictor (predictive model) to each node of the tree; we call the resulting object a  {\em  tree of predictors}.  The (locally) optimal tree of predictors is derived recursively; each step involves {\em jointly optimizing} the split of the terminal nodes of the previous tree and the choice of learner (from among a given set of base learners) and training set -- hence predictor -- for each set in the split.  The features of a new instance determine a unique path through the optimal  tree of predictors; the final prediction aggregates the predictions of the predictors along this path.  Thus, our approach uses base learners to create  complex learners that are matched to the characteristics of the data set while avoiding overfitting.  We establish loss bounds for the final predictor in terms of the Rademacher complexity of the base learners.  We report the results of  a number of experiments  on a variety of datasets, showing that our approach provides statistically significant improvements over a wide variety of state-of-the-art machine learning algorithms, including various ensemble learning methods.     
    \end{abstract}
    

    %
    \IEEEpeerreviewmaketitle

    \section{Introduction}
    %
    %
    %
    %
    \IEEEPARstart{E}{nsemble} methods \cite{Ensemble_General,Ensemble_Cem,Ensemble_HMM,Ensemble_Luca} are general techniques in machine learning that combine several learners; these techniques include  bagging \cite{ref1,ref4,ref5}, boosting \cite{Ensemble_Boosting,ref2} and stacking \cite{ref3}. Ensemble methods frequently improve predictive performance. We describe a novel approach to ensemble learning that chooses the learners  to be used, the way in which these learners should be trained to create predictors (predictive models), and the way in which the predictions of these predictors should be combined according to the {\em features} of a new instance for which a prediction is desired.   By {\em jointly} deciding which learners and training sets to be used, we provide a novel method to grow complex predictors; by deciding how to aggregate these predictors we control overfitting.  As a result, we obtain substantially improved predictive  performance. 
    
    Our proposed model, ToPs (Trees of Predictors), differs from existing methods in that it constructs and applies {\em different} predictive models to {\em different} subsets of the feature space.  Our approach has something in common with  tree-based approaches (e.g. Random forest, Tree-bagging, CART, etc.) in that we successively split the feature space.  However, while tree-based approaches create successive splits of the feature space in order to maximize {\em homogeneity} of each split {\em with respect to labels}, ToPs creates successive splits of the feature space in order to maximize  {\em predictive accuracy} of each split {\em with respect to a constructed predictive model}.  To this end, ToPs creates a tree of subsets of the feature space and associates a predictive model to each such subset -- i.e., to each node of the tree.  To decide whether to  split a given node, ToPs uses a feature to create a tentative split, then chooses a learner and a training set to create a predictive model for each set in the split, and searches for the feature, the learner and the training set that maximizes the predictive accuracy (minimizes the prediction error).  ToPs continues this process recursively until no further improvement is possible.  
    
    \begin{figure}[t!]
        \centering
        \includegraphics[width=0.48\textwidth]{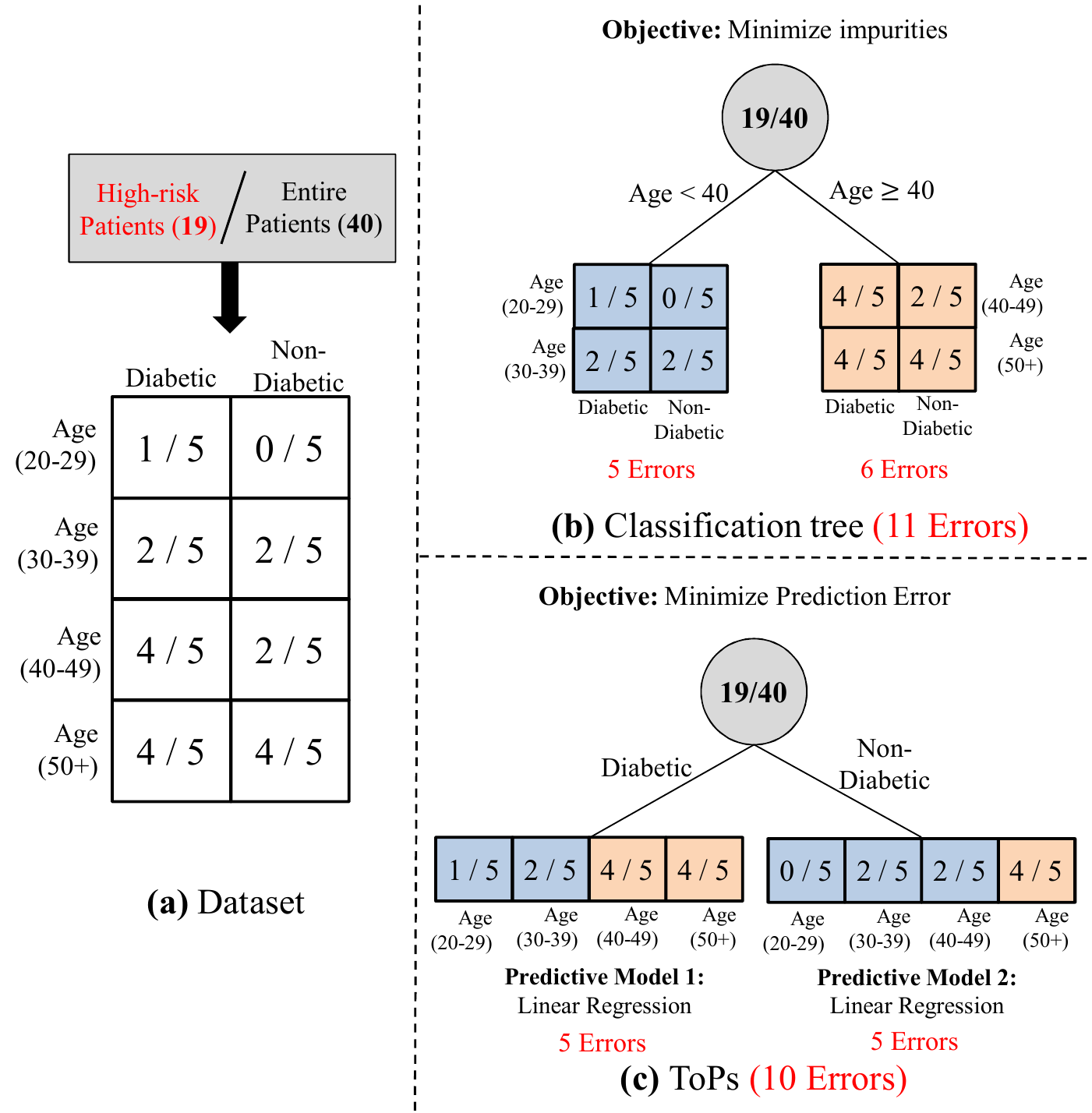}
        \caption{Toy example. (a) Dataset, (b) Classification tree, (c) ToPs}
        \label{figure:example}
    \end{figure}
    
    A simple toy example, illustrated in Fig. \ref{figure:example}, may help to illustrate how ToPs works and why it improves on other tree-based methods. We consider a classification problem: making binary predictions of  hypertension in a patient population.  We assume two  features: Diabetic or Non-diabetic and Age Range 20-29, 30-39, 40-49, 50+, so that there are 8 categories of patients.  We assume the data is as shown in Fig. \ref{figure:example} (a): there are 5 patients in each category; 1 patient who is Diabetic and in the Age Range 20-29 has hypertension, etc. We first consider a simple classification tree. Such a tree first selects a single feature and threshold to split the population into two groups so as to maximize the purity of labels. In this case the best split uses Age Range as the splitting feature and partitions patients into those in the Age Ranges 20-29 or 30-39  and those in the Age Ranges 40-49 or 50+.  No further splitting improves the purity of labels so we are left with the tree shown in Fig. \ref{figure:example} (b). The resulting predictive model  predicts that those in the Age Ranges 20-29 or 30-39 will not have hypertension and those in the Age Ranges 40-49 or 50+ will have hypertension; this model makes 11 prediction errors. We now consider an instantiation of ToPs which uses as base learner a linear regression (to produce a probability of hypertension) followed by thresholding at 0.5. (Recall that we are treating this as a classification problem.) The best split for this instantiation of ToPs uses Diabetic/Non-diabetic as the splitting feature (leading to the split shown in Fig. \ref{figure:example} (c)) but creates different predictive models in the two halves of the split: in the Non-diabetic half of the split, the model predicts that patients in the Age Ranges 20-29, 30-39 and 40-49 will not have  hypertension and that patients in the Age Range 50+ will have hypertension; in the Diabetic half of the split the model predicts that patients in the Age Ranges 20-29 and 30-39 will not have  hypertension and that patients in the Age Ranges 40-49 and 50+ will have hypertension. (After this split, no further splits using this single base learner improve the prediction accuracy.) This  model makes only 10 errors (less than the number of errors made by the classification tree). Note that the tree produced by ToPs is completely different from the classification tree, that the predictive models and predictions produced by ToPs are different from those produced by the classification tree, and that the predictions produced by ToPs within a single terminal node are not uniform. In this case, ToPs performs better than the classification tree because it "understands" that the effect of age on the risk of hypertension is different for patients who are Diabetic and for patients who are Non-diabetic.
    
    Although this toy example may seem artificial, it exemplifies what happens when we apply ToPs to a real dataset.  For example, one of our experiments is survival prediction of patients who are wait-listed for a heart transplant.  (For more discussion, see Section \ref{sect:experiments} and \ref{sect:discussion}.)  In that setting features $X$ are patient characteristics and labels $Y$  are survival times.  The data set $\data$ consists of records of actual patients; a single data point $(x^t,y^t)$  records that a patient with features $x^t$ survived for time $y^t$.  The construction of our algorithm demonstrates that the {\em best predictor} of survival for males is {\em different} from the {\em best predictor} of survival for females.  As a result, predictions of survival for a male and a female with otherwise similar features may be quite different - because the features that influence survival have different importance and interact differently for males and females.  Using a gender-specific predictor leads to significant improvement in prediction accuracy.
    
    In what follows, Section \ref{sect:comparisons} highlights the differences between our method and related machine learning methods. Section \ref{sect:method} provides a full description of our method.  Section \ref{sect:bounds} derives loss bounds.  Section \ref{sect:experiments} compares the performance of our method with that of many other methods on a variety of datasets,  demonstrating that our method provides substantial and statistically significant improvement.  Section \ref{sect:discussion} details the operation of our algorithm for one of the datasets to further illustrate how why our method works.   Section \ref{sect:conclusion} concludes. Proofs are in the Appendix (at the end of the manuscript);  parameters of the experiments and additional figures and discussion can be found in the Supplementary Materials.

    \begin{table*}[t!]
        \caption{Comparison with existing methods}    
        \label{table:related}
        \centering
        \begin{tabular}{|M|M|L|L|}
        \toprule
            \textbf{Methods} & \textbf{Sub-categories} & \textbf{Approach of existing methods} & \textbf{Approach of ToPs}  \\ \midrule
            \textbf{Ensemble Methods} & \textbf{Bagging, Boosting, Stacking} & Construct multiple predictive models (with single learning algorithm) using randomly selected training sets. & Construct multiple predictive models using multiple learning algorithms and optimal  training sets.\\  \cmidrule{3-4}
            & & Only the predictive models are optimized to minimize the loss. & Training sets and the assigned predictive model for each training set are jointly optimized to minimize the loss \\ \midrule
            \textbf{Tree-based Methods} & \textbf{Decision Tree, Regression tree} & Grow a tree by choosing the split that minimizes the impurities of labels in each node.  & Grow a tree by choosing the split that minimizes the prediction error of the best predictive models. \\  \cmidrule{3-4}
            &&Within a single terminal node all the predictions are uniform (identical) .&Within a single terminal node, the final predictive model is uniform but the predictions can be different. \\ \midrule
            \textbf{Non-parametric Regression} & \textbf{Gaussian Process, Kernel Regression} & Construct a non-parametric predictive model with pre-determined kernels using the entire training set.  & Construct  non-parametric predictive models by jointly optimizing the training sets and the best predictive models. \\ \cmidrule{3-4}
            &&Similarities between data points are fixed functions of the features. & Similarities between data points are determined by the prediction errors. \\ \midrule
            \textbf{Model trees} & \textbf{[12], [13], [14], [15], [16]} & Construct the tree by jointly optimizing a (single) learning algorithm and the splits & Construct the tree by jointly optimizing multiple learning algorithms and the splits \\  \cmidrule{3-4}
            && Only convex loss functions are allowed. & Arbitrary loss functions such as AUC are allowed. \\ \cmidrule{3-4}
            && The final prediction for a new instance depends only on the terminal node to which the feature of the instance belongs. & The final prediction is a weighted average of the predictions along all the nodes to which the feature of the instance belong (the path to the terminal node). \\ \bottomrule
        \end{tabular}
    \end{table*}
    
    \section{Methodological Comparisons}\label{sect:comparisons}  
    ToPs is most naturally compared with four previous bodies of work: model trees, other tree-based methods, ensemble methods, and non-parametric regression. Table \ref{table:related} summarizes the comparisons with existing methods. The optimization equations of existing works are also compared in the Appendix Table \ref{table:related_work}.
    
        \begin{figure}[t!]
        \centering
        \includegraphics[width=0.5\textwidth]{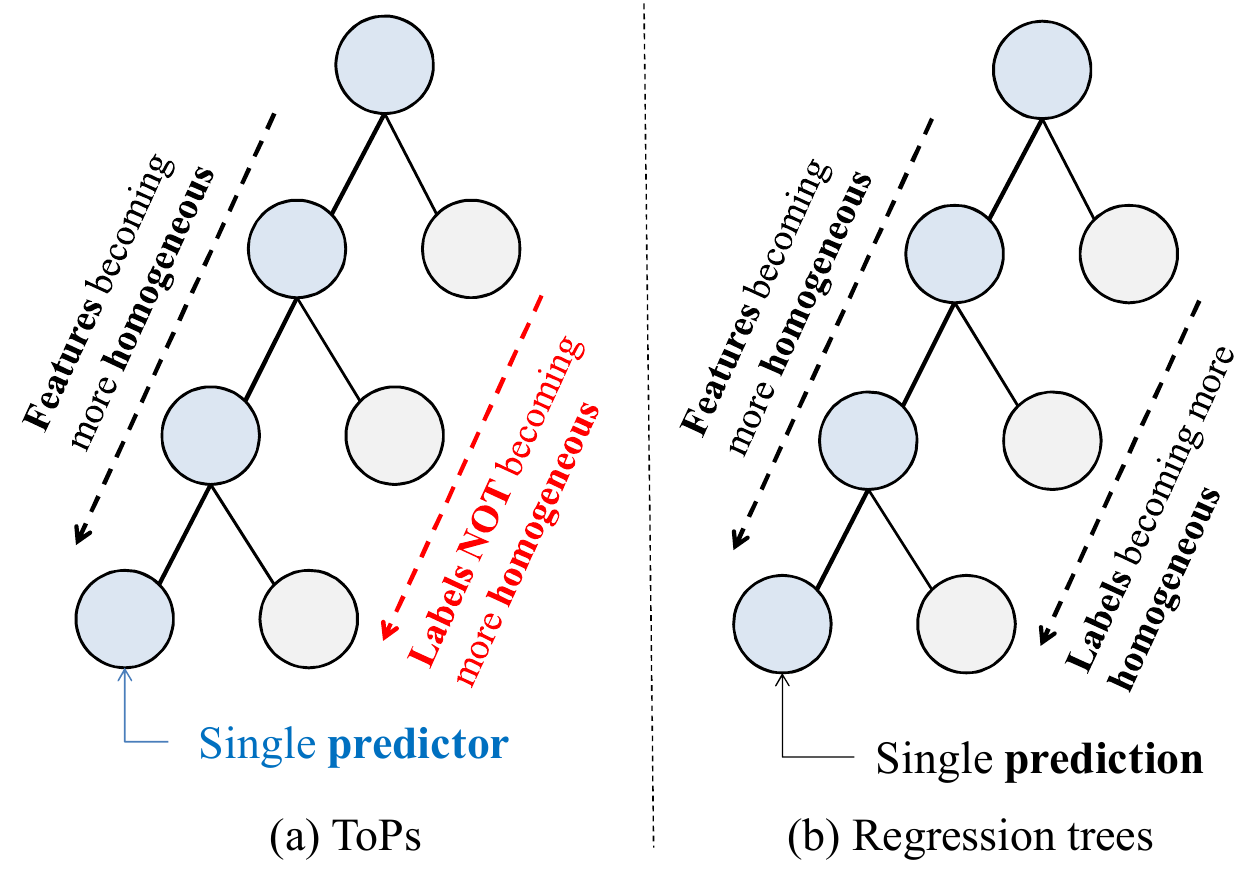}
        \caption{The main differences between ToPs and regression trees. (a) ToPs, (b) regression trees}
        \label{figure:novelty}
    \end{figure}
    
    \subsection{Model trees}\label{subsect:model_trees}
    There are similarities between ToPs and model trees but also very substantial differences.  The initial papers on model trees  \cite{ref12,ref13}  construct a tree using a splitting criterion that depends on labels but not on predictions, then prunes the tree, using linear regression at a node to replace subtrees from that node.  \cite{ref14} constructs the tree in the same way but allows for more general learners when replacing subtrees from a node. \cite{ref15,ref16} split the tree by jointly optimizing a linear predictive model and the splits that minimize the loss or maximize the statistical differences between two splits. 
    
    ToPs operates  differently from all of these: it constructs the tree using a splitting criterion that depends {\em both} on labels {\em and} on predictions of its base learners,  it does not prune its tree nor does it replace any subtree with a predictive model.    
    
    In ToPs, the potential split of a node is evaluated by training {\em each base learner} on the current node and on all parent nodes (not just the current node, as in \cite{ref15,ref16}) and choosing the split {\em and} the predictor that yield the best overall performance.  (Keep in mind that ToPs is a general framework that allows for an arbitrary set of base learners.)  This approach is important for several reasons: (1) It avoids the problem of small training sets and reduces overfitting. (2) It allows splits even when only one side of the split yields an improved performance. (3) It allows for choosing the base learner among multiple learners that best capture the importance of features and the interactions among features.  ToPs  constructs the final prediction as a weighted average of predictions along the path (with weights  determined by linear regression).  Weighted averaging is important because it smooths the prediction, from the least biased model to the most biased model. Determining the weights by linear regression is important because it is an optimizing procedure, and so the weights  depend on the data and on the accuracy of the predictors constructed.  (\cite{ref16} also smooths by weighted averaging but  chooses the weights according to the number of samples and number of features.  This makes sense only with a single base learner and hence a single model complexity.  With multiple  models of different model complexities, it is impossible to find the appropriate aggregation weights using only  the number of features and number of samples).     Finally, the construction of ToPs allows for arbitrary loss functions, including loss  functions such as AUC that are not sample means.  This is important because loss functions such as AUC are especially appropriate for certain applications; e.g. survival in the heart transplant dataset.

    \subsection{Other tree-based methods} Decisions trees, regression trees \cite{Sup1}, tree bagging, and random forest \cite{ref1}  follow a recursive procedure, growing a tree by using features to create tentative splits and labels to choose among the tentative splits. Eventually, these methods create a final partition (the terminal nodes of the tree) and make a single uniform prediction within each set of this partition.  Some of these methods construct multiple trees and hence multiple partitions and aggregate the predictions arising from each partition.  Our method also follows a recursive procedure, growing a tree by using features to create tentative splits, but we then use features {\em and} labels {\em and} predictors to choose the optimal split and associated predictors.  Eventually, we produce a (locally) optimal tree of predictors.  The final prediction for a given new instance is computed by aggregating the predictions along a path in this tree of predictors.  A crucial difference between other tree-based methods and ToPs is the treatment of instances that give rise to the same paths: in other tree-based methods, such instances will be assigned {\em the same prediction}; in ToPs, such instances will be assigned {\em the same predictor} but may be assigned {\em very different predictions}.   Other tree-based models have the property that within a single terminal node all the predictions are uniform; ToPs has the property that within a single terminal node the predictor is uniform but the predictions can be different.  See again the illustrative example above and Fig. \ref{figure:novelty}.  (ToPs also has something - but less - in common with hierarchical logistic regression and hierarchical trees; in view of space constraints we defer the discussion to the Supplementary Materials.)
    
    \subsection {Ensemble methods} 
    Bagging \cite{Sup3}, boosting \cite{ref2,ref7,ref9,ref11}, stacking \cite{ref3,Sup9} construct multiple predictive models using different training sets and then aggregate the predictions of these models according to endogenously determined weights.  Bagging methods use a single base learner and choose random training sets (ignoring both features and labels).  Boosting methods use a single base learner and choose a sequence of training sets to create a sequence of predictive models; the sequence of training sets is created recursively according to random draws from the entire training set but weighted by the errors of the previous predictive model.  Stacking uses multiple base learners with a single training set to construct multiple predictive models and then aggregates the predictions of these multiple predictive models.  Our method uses a recursive construction to construct a locally optimal tree of predictors, using both multiple base learners and multiple training sets and constructs optimal weights to aggregate the predictions of these predictors.
    
    \subsection {Non-parametric regressions} Kernel regression \cite{Sup10,Sup11} and Gaussian process regression \cite{Sup12} have in common that given a feature $x$, they choose a training set and use that set to determine the coefficients of a linear learner/model to predict the label $y$.  Kernel regression begins with a parametric family of kernels $\{K_\theta\}$.  For a specific vector $\theta$  of parameters and a specific bandwidth $b$, the regression considers the set $B$ of data points $(x^t,y^t)$ whose feature $x^t$ are within the specified bandwidth $b$ of $x$; the predicted value of the corresponding label $y$ is the weighted sum $\sum_B K_\theta(x,x^t) \, y^t$.  The optimal parameter $\theta^*$ and bandwidth $b^*$ can be set by training, typically using least squared error as the optimization criterion.  Gaussian process regression begins by assuming a Gaussian form for the kernel but with unknown mean and variance.  It first uses a maximum likelihood estimator based on the entire data set to determine the mean and variance (and perhaps a bandwidth), and then uses that the Gaussian kernel to carry out a kernel regression.  In both kernel regression and Gaussian process regression, the prediction for a new instance is formed by aggregating the labels associated to nearby features.  In our method, the prediction for a new instance is formed by carefully aggregating the predictions of a carefully constructed family of predictors.

    \section{ToPs}\label{sect:method}

    We work in a supervised setting so data is presented as a pair $(x,y)$ consisting of a feature and a label.  We are presented with a (finite) dataset $\data = \{(x^t,y^t)\}$, assumed to be drawn iid from the true distribution $\dist$, and seek to learn a model that predicts, for a new instance drawn from $\dist$ and for which we observe the feature vector $x$, the true label $y$.  We assume the space of features is $X = X_1 \times \cdots X_d$;  if  $x \in X$ then $x_i$ is the $i$-th feature.  Some features are categorical, others are continuous.  For convenience (and without much loss of generality) we assume categorical features are binary and represented as $0,1$ and that continuous features are normalized between $0,1$; hence $X_i \subset [0,1]$ for every $i$ and $X \subset [0,1]^d$.  We also take as given a set $Y \subset \R$ of {\em labels}.   For $Z \subset X$, a {\em predictor} ({\em predictive model}) on $Z$ is a map $h: Z \to Y$ or $h: Z \to \mathbf{R}$; we interpret  $h(z)$ as the predicted label or the expectation of the predicted label, given the feature $z$.  (We often suppress $Z$ when $Z = X$ or when $Z$ is understood.)  If $h_1: Z_1 \to Y, h_2 : Z_2 \to Y$ are predictors and $Z_1 \cap Z_2 = \emptyset$ we define $h_1 \vee h_2: Z_1 \cup Z_2 \to Y$ by $h_1 \vee h_2(z) = h_i(z)$ if $z \in Z_i$. All the notations are summarized in the nomenclature table (Table \ref{table:nomenclature}).

    \begin{table}[t!]
        \caption{Nomenclature table}        
        \label{table:nomenclature}
        \centering
        \begin{tabular}{|p{1cm}|p{7cm}|}
            \toprule
            \textbf{Symbol} & \textbf{Explanation} \\ \midrule
            $\mathcal{A}$ & A set of algorithms \\
            $A$ & An algorithm ($\in \mathcal{A}$) \\
            $C$ & A node \\
            $C^\uparrow$ & Set consisting of node $C$ and all its predecessors \\
            $\bar{C}(x)$ & The unique terminal node that contains $x$ \\
            $C^+, C^-$ & Divided subsets (Similarly for $h, A, S$ and $V$)\\
            $D$ &  Dataset $=\{(x^t, y^t)\}$ \\
            $\mathcal{D}$ & True distribution \\
            $h_C$ & Predictive model assigned to $C$ \\
            $H(x)$ & Overall prediction assigned to $x$ by the overall predictor $H$ \\
            $L(h,Z)$ & The loss with a predictor $h$ and the set $Z$ \\
            $M$ & Number of algorithms \\
            $N$ & Number of samples \\
            $\Pi(x)$ & Path from the initial node to the terminal node with $x$ \\
            $S$ & Training set \\
            $\tau$ & Threshold to divide nodes \\
            $\mathcal{T}$ & A tree \\
            $\bar{\mathcal{T}}$ & Set of  terminal nodes \\
            $V^1, V^2$ & Two validation sets \\
            $w^*(C,\Pi)$ & Weights of $C$ on the path $\Pi$ \\
            $X$ & Feature space \\
            $(x^t, y^t)$ & Instance. $x^t$: feature, $y^t$: label\\
            $x_i$ & $i$-th feature \\
            $Y$ & Label set \\
            $Z$ & Subspace of the feature space \\
            \bottomrule
        \end{tabular}
    \end{table}

    We take as given a finite family $\algs = \{A_1, \ldots, A_M\}$ of {\em algorithms} or {\em base learners}. We interpret an algorithm as including the parameters of that algorithm (if any); thus Random Forest with 100 trees is a different algorithm than Random Forest with 200 trees.  Given an algorithm $A \in \algs$ and a set $E \subset \data$ to be used to train $A$, we write $A(E)$ for the resulting predictor.  If $\mathcal E$ is a family of subsets of 
    $\data$ then we write $\algs({\mathcal E}) = \{A(E) : A \in \algs, E \in {\mathcal E} \}$  
    for the set of predictors that can arise from training some algorithm in $\algs$ on some set in $\mathcal E$.
    
    A {\em  tree of predictors} is a pair $(\tree, \{h_C\})$ consisting of a (finite) family $\tree$ of non-empty subsets of $X$ together with an assignment $C \mapsto h_C$ of a predictor $h_C$ on $C$ to each element  $C \in \tree$ such that: 
    \begin{enumerate}[(i)]
        \item $X \in \tree$. 
        
        \item With respect to the ordering induced by set inclusion, $\tree$ forms a tree; i.e., $\tree$ is partially ordered in such a way that each  $C \in \tree$, $C \not= X$ has a unique immediate predecessor.  As usual, we refer to the elements of $\tree$ as {\em nodes}.  Note that $X$ is the initial node.  
        
        \item If $C \in \tree$ is not a terminal node then the set $C^{\rm succ}$ of immediate successors of $C$ is a partition of $C$. 
        
        \item For each node $C \in \tree$ there is an algorithm $A_C \in \algs$ and a node $C^* \in \tree$ such that $C \subset C^*$ (so that either $C = C^*$ or $C^*$ precedes $C$ in $\tree$) and $h_C = A_C(C^*)$ is the predictor formed by training the algorithm $A_C$ on the training set $C^*$. \footnote{In principle, the predictive model $h_C$ and/or the algorithm $A_C$ and the node $C^*$ might not be unique.  In that case, we can  choose randomly among the possibilities. Because this seems an unusual situation, we shall ignore it and similar indeterminacies that may occur at other points of the construction.}
        
    \end{enumerate}
    It follows from these requirements that for any two nodes $C_1, C_2 \in \tree$ exactly one of the following must hold: $C_1 \subset C_2$ or $C_2 \subset C_1$ or $C_1 \cap C_2 = \emptyset$.  It also follows that the set of terminal nodes forms a partition of $X$.  For any node $C$ we write $C^\uparrow$ for the set consisting of $C$ and all its predecessors.  For $ \in X$ we write $\overline{C}(x)$ for the unique terminal node that contains $x$ and $\Pi(x) = \Pi(\overline{C}(x))$ for the unique path in $\tree$ from the initial node $X$ to the terminal node $\overline{C}(x)$.  We write $\overline{\tree}$ for the set of all terminal nodes.
    
    \begin{figure}[t!]
        \centering
        \includegraphics[width=0.5\textwidth]{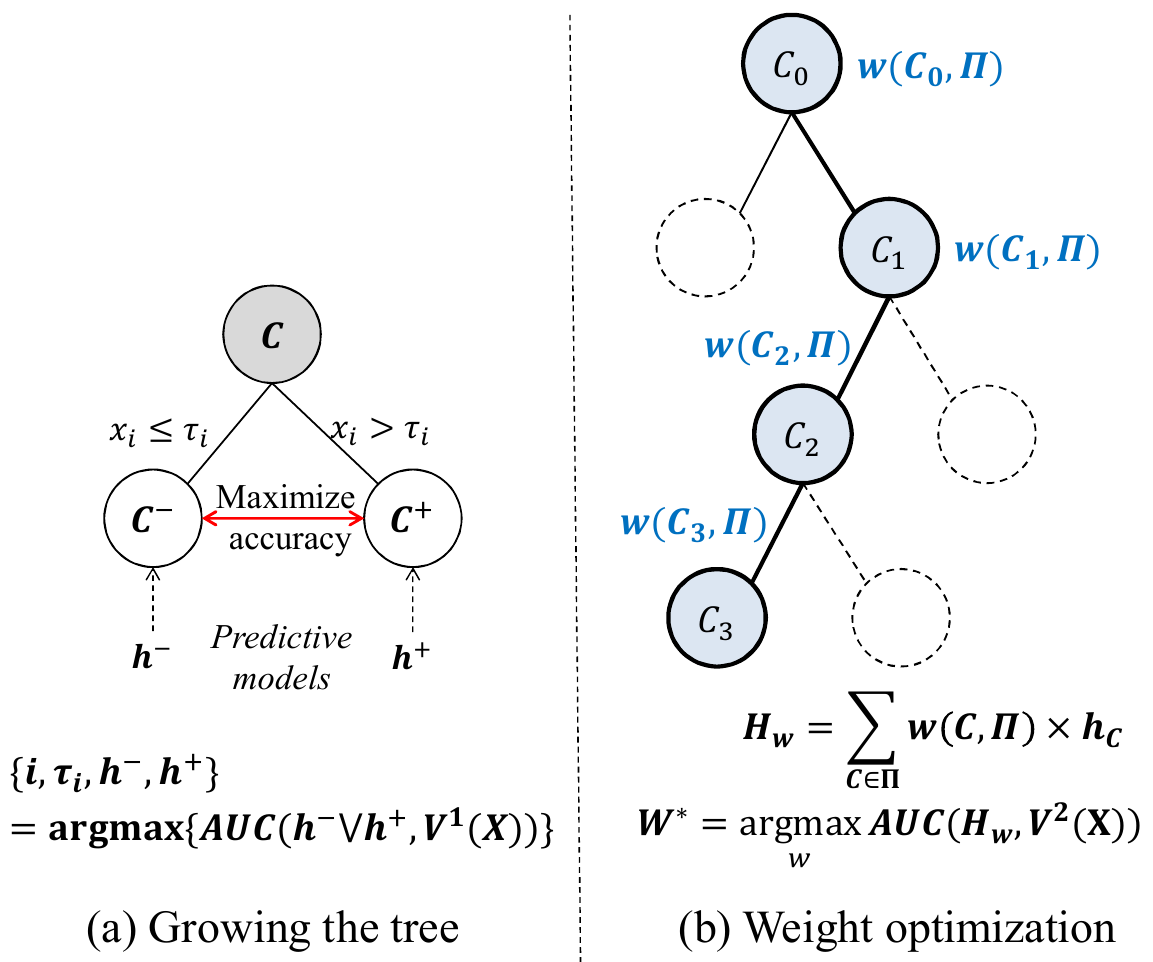}
        \caption{Illustrations of ToPs (a) growing the tree, (b) weight optimization}
        \label{figure:notation}
    \end{figure}
    
    We fix a partition of the given dataset  $\data = S \cup V^1 \cup V^2$; we view $S$ as the (global) training set and $V^1, V^2$ as   (global) validation sets. In practice, the partition of $\data$ into training and validation sets will be chosen randomly.  Given a set $C \subset X$ of feature vectors and a subset $Z \subset \data$, write $Z(C) = \{(x^t,y^t) \in Z: x^t \in C\}$.  
    
    We express performance in terms of loss.  For many problems, an appropriate measure of performance is the Area Under the (receiver operating characteristic) Curve (AUC); for a given set of data $Z \subset D$ and  predictor $h: Z(X) \to Y$ the loss is $L(h,Z) = 1 - AUC$.  For other problems, an appropriate measure of loss is the sample mean error $L(h,Z) = \frac{1}{|Z|} \sum_{(x,y) \in Z}  |h(x) - y|$.  For our general model, we allow for an arbitrary loss function. Given disjoint sets $Z_1, \ldots, Z_N$ and predictors $h_n$ on $Z_n$, we measure the {\em total (joint) loss} as $L(\bigvee h_n, \bigcup Z_n)$.  Note that the  loss function need not be additive, so in general $L(\bigvee h_n, \bigcup Z_n) \not= \sum L(h_n, Z_n)$
    
    \subsection{Growing the  (Locally) Optimal  Tree of Predictors} 
    
    We begin with the (trivial) tree of predictors $(\{X\}, \{h_X\})$ where $h_X$ is  the predictor  $h \in \alg(\{S\})$ that minimizes the   loss $L(h,V^1(X))$. Note that we  train $h$ globally -- on the entire training set  - and evaluate/validate it globally - on the entire (first) validation set - because the initial node consists of the entire feature space.  We now grow the tree of predictors by a recursive splitting process, which is illustrated in Fig. \ref{figure:notation} (a).  Fix the tree of predictors $(\tree, \{h_C\})$ constructed to this point.   For each terminal node $C \in \tree$ with its associated predictor $h_C$, choose a feature $i$ and a threshold $\tau_i \in [0,1]$.  (In practice, for binary variables, the threshold is set at 0.5. For continuous variables, the thresholds are set to delineate percentile boundaries at 10\% and 90\%, with increments of 10\%.)  Write 
    \begin{align*}
    C^-(\tau_i) &= \{x \in C : x_i < \tau_i \} \\
    C^+(\tau_i) &= \{x \in C : x_i \geq \tau_i \} 
    \end{align*}
    Evidently $C^-(\tau_i), C^+(\tau_i)$ are disjoint and $C = C^-(\tau_i) \cup C^+(\tau_i)$, so we are splitting $C$ according to the feature $i$ and the threshold $\tau_i$.  Note that $C^-(\tau_i) = \emptyset$ if $\tau_i = 0$; in this case we are not {\em properly} splitting $C$.  For each of $C^-(\tau_i) , C^+(\tau_i)$ we choose  predictors $h^- \in \alg(C^-(\tau_i)^\uparrow), h^+ \in \alg(C^+(\tau_i)^\uparrow)$.  (That is, we choose predictors that arise from one of the learners $A \in \alg$, trained on {\em some} node that (weakly) precedes the given node.) We then choose the feature $i^*$, the threshold $\tau_i^*$, and the predictors  $h_{C^-(\tau_i^*)}, h_{C^+(\tau_i^*)}$ to minimize the total loss on $V^1$; i.e. we choose them to solve the minimization problem 
    \begin{align*}
    \arg\min_{i, \tau_i,  h^-, h^+}  L\Bigl( h^- \vee h^+, V^1 \bigl(C^-(\tau_i) \bigr) \cup V^1 \bigl(C^+(\tau_i) \bigr) \Bigr)  
    \end{align*}
    subject to the requirement that this total loss should be strictly less than $L(h_C, V^1(C))$.  Note that because choosing the threshold $\tau_i = 0$ does not properly split $C$, the loss $L(h_C, V^1(C))$ can always be achieved by setting $\tau_i = 0$ and $h^+ = h_C$.   If there is no proper splitting that yields a total loss smaller than $L(h_C, V^1(C))$ we do not split this node.  This yields a new tree of predictors.  We stop the entire process when no terminal node is further split; we refer to the final object as the {\em locally optimal tree of predictors}. (We use the adjective "locally" because we have restricted the splitting to be by a single feature and a single threshold and because the optimization process employs a greedy algorithm -- it does not look ahead.)

    \begin{algorithm}[tb]
        \caption{Growing the  Optimal  Tree of Predictors}
        \label{alg:Algo1}
        \begin{algorithmic}
            \STATE {\bf Input:} Feature space $X$, a set of algorithms $\alg$, training set $S$, the first validation set $V^1$
            \STATE {\bf First step:}
            \STATE Initial tree of predictors $= (X,h_X)$, 
            \STATE where $h_X = \arg\min_{A \in \alg} L(A(S),V^1)$
            \STATE {\bf Recursive step:} 
            \STATE {\bf Input:} Current tree of predictors ($(\tree, \{ h_C \}$)
            \STATE {\bf For} each terminal node $C \in \mathcal{T}$
            \STATE \hspace{2 mm} {\bf For} a feature $i$ and a threshold $\tau_i \in [0,1]$.
            \STATE \hspace{4 mm} Set $C^-(\tau_i) = \{x \in C :x_i < \tau_i \}$
            \STATE \hspace{9.5 mm}  $C^+(\tau_i) = \{x \in C :x_i \geq \tau_i \}$
            \STATE \hspace{4 mm} Then,
            \STATE \hspace{4 mm} $\{i^*,\tau_i^*,h_{C^-(\tau_i^*)}, h_{C^+(\tau_i^*)}\} = $
            \STATE \hspace{4 mm} ${\rm arg}\min L\Bigl( h^- \vee h^+, V^1 \bigl(C^-(\tau_i) \bigr) \cup V^1 \bigl(C^+(\tau_i) \bigr) \Bigr) $
            \STATE \hspace{4 mm} where $h^- \in \alg(C^-(\tau_i)^{\uparrow}), h^+ \in \alg(C^-(\tau_i)^{\uparrow})$
            \STATE \hspace{2 mm} {\bf End For}
            \STATE {\bf End For}
            \STATE {\bf Stopping criterion: }
            \STATE $L(h_C,V^1(C)) \leq $
            \STATE $\min L\Bigl( h^- \vee h^+, V^1 \bigl(C^-(\tau_i) \bigr) \cup V^1 \bigl(C^+(\tau_i) \bigr) \Bigr)$
            \STATE {\bf Output:} Locally optimal tree of predictors ($\tree, \{ h_C \})$
            
        \end{algorithmic}
    \end{algorithm}

    \subsection{Weights on the Path}
    Fix the locally optimal tree of predictors$(\tree, \{h_C\})$ and a terminal node $\overline{C}$; let $\Pi$ be the path from $X$ to $\overline{C}$.  We consider vectors  ${\mathbf w} = \bigl( w(\Pi, C)\bigr)_{C \in \Pi}$ of non-negative weights summing to one; for each such weight vector we form the predictor $H_{\mathbf w} = \sum_{C \in \Pi} w(C,\Pi) \, h_{C}$.  We choose the weight vector ${\mathbf w^*}(\Pi)$  that minimizes the empirical loss of $H_{\bf w}$ on the second validation set $V^2(\overline{C})$; i.e.
    $$
    {\mathbf w^*}(\Pi) = {\rm arg}\min_{\mathbf w} L\bigl( H_{\mathbf w} , V^2(\overline{C}) \bigr)
    $$
    By definition, the weights depend on the path and not just on the node; the weights assigned to a node $C$ along different paths may be different.  This procedure is illustrated in Fig. \ref{figure:notation} (b).    
    
    \begin{algorithm}[tb]
        \caption{Weights on the Path}
        \label{alg:Algo2}
        \begin{algorithmic}
            \STATE {\bf Input:} Locally optimal tree of predictors ($\tree,\{ h_C \})$,  second validation set $V^2$
            \STATE {\bf For} each terminal node $\overline{C}$ and the corresponding path $\Pi$ from $X$ to $\overline{C}$
            \STATE \hspace{2 mm} For each weight vector ${\mathbf w} = (w_C)$, 
            \STATE \hspace{2 mm} Define $H_{\mathbf w} = \sum_{C \in\Pi} w_C h_C$. Then,
            \STATE \hspace{2 mm} $ {\mathbf w^*}(\Pi) =  \bigl( w^*(\Pi,C) \bigr) = \arg\min  L\bigl(H_{\mathbf w},V^2(\overline{C})\bigr)  $
            \STATE {\bf End For}
            \STATE {\bf Output:} Optimized weights ${\mathbf w^*}(\Pi)$ for each terminal node $\overline{C}$ and corresponding path $\Pi$
            
        \end{algorithmic}
    \end{algorithm}
    
    \subsection{Overall Predictor}
    Given the locally optimal tree of predictors $(\tree, \{h_C\})$ and the optimal weights $\{{\mathbf w^*}(\Pi)\}$ for each path, we define the overall predictor $H: X \to \R$ as follows:  Given a feature $x$, we define
    $$
    H(x) =  \sum_{C \in \Pi(x)}  {\mathbf w^*}(\Pi(x),C) \, h_C (x)
    $$
    That is, we compute the weighted sum of all predictions along the path $\Pi(x)$ from the initial node  to the  terminal node that contains $x$.  
    
    Note that we construct predictors by training algorithms on (subsets of) $S$ but we construct the locally optimal tree by minimizing losses with respect to (subsets of) the validation set $V^1$; this avoids overfitting to the training set $S$.  We then construct the weights,  hence the overall predictor, by minimizing losses with respect to (subsets of) the validation set $V^2$; this avoids overfitting to the validation set $V^1$. Fig. \ref{figure:notation} (b) illustrates the procedures. The pseudo-codes of the entire ToPs algorithms are in Algorithm \ref{alg:Algo1}, \ref{alg:Algo2} and \ref{alg:Algo3}.
    
    \begin{algorithm}[tb]
        \caption{Overall Predictor}
        \label{alg:Algo3}
        \begin{algorithmic}
            \STATE {\bf Input:} Locally optimal tree of predictions $(\tree,\{ h_C \})$, optimized weights $\{{\mathbf w^*}(\Pi)\}$, and  testing set $T$
            \STATE {\bf Given a feature vector $x$}, 
            \STATE Find unique path $\Pi(x)$ from $X$ to terminal node  containing $x$;
            \STATE {\bf Then}
            \STATE $H(x)= \sum_{C \in \Pi(x)} {\mathbf w^*}(\Pi, C) \, h_C(x) $
            \STATE {\bf Output:} The final prediction $H(x)$
            
        \end{algorithmic}
    \end{algorithm}
    
    \subsection{Instantiations}
    It is important to keep in mind that we take as given a family $\alg$ of base learners (algorithms).  Our method is independent of the particular family we use, but of course, the final overall predictor is not.  We use ToPs to refer to our general method and to ToPs/$\alg$ to refer to a particular instantiation of the method, built on top of the family $\alg$ of base learners; e.g., ToPs/LR is the instantiation build on Linear Regression as the sole base learner.    In our experiments, we compare the performance of two different instantiations of ToPs.

    \subsection{Computational Complexity}
    The computational complexity of constructing the tree of predictors is $\sum_{i=1}^M \mathcal{O}(N^2  D \times T_i(N,D))$ where $N$ is the number of instances, $D$ is the number of features, $M$ is the number of algorithms and $T_i(N,D)$ is the computational complexity of the $i$-th algorithm. (The proof is in the Appendix.) For instance, the computational complexity of ToPs/LR is $\mathcal{O}(N^3  D^3)$. The computational complexity of finding the weights is low in comparison with constructing the tree of predictors because it is just a linear regression. In all our simulations, the entire training time was less than 13 hours using an Intel 3.2GHz i7 CPU with 32GB RAM. (Details are in the Supplementary Materials.)  Testing can be done in real-time without any delay.

    \section{Loss Bounds}\label{sect:bounds}  
    In this Section, we show how the Rademacher complexity of the base learners can be used to provide loss bounds for our overall predictor.  The loss bound we establish is an application to our framework of the Rademacher Complexity Theorem. The importance of the loss bound we establish is that, even though we use multiple learning algorithms with multiple clusters (produced by divisions/splits), the loss can be bounded by the loss of the most complicated single learner. Hence, the sample complexity of ToPs is at most the sample complexity of the most complex single learner.
     
     Throughout this section we assume the loss function is the sample mean of individual losses: $L(h,Z) = \frac{1}{|Z|} \sum_{z \in Z} \ell(h,z)$), and that the individual loss is a convex function of the difference $\ell(h,z) = \hat{\ell}(h(x)-z)$ 
     where $\hat{\ell}$ is convex. Note that mean error and mean squared error satisfy these assumptions but that other loss functions such as $1-AUC$ (which is the natural loss function in the setting of our heart transplant experiment described in Section IV-D) does not. Recall that for each node $C$,  $A_C$ is the base learner used to construct the predictor $h_C$ associated to the node $C$ and that for a terminal node $\overline{C}$ we write $\Pi(\overline{C})$ for the path from $X$ to $\overline{C}$.  We first present an error bound for each individual terminal node and then derive an overall error bound.  (Proofs are in the Appendix.)  We write  
    $ \mathcal{R}(A_C,S(\bar{C}))$ for the Rademacher complexity of $A_C$ with respect to the  portion of the training set $S(\overline{C})$ and 
    $\mathbb{E}_{\mathcal{D}_{\overline{C}}} [L(H)]$ for the expected loss of the overall predictor $H$ with respect to the true distribution when features are restricted to lie in $\overline{C}$ and $\mathbb{E}_{\mathcal{D}} [L(H)]$ for the expected loss of the overall predictor $H$ with respect to the true distribution.

    \begin{theorem}  Let $H$ be the overall predictor and let $\overline{C}$ be a terminal node of the locally optimal tree.  For each  $\delta > 0$, with probability at least $1-\delta$ we have
        \begin{align*}
        \mathbb{E}_{\mathcal{D}_{\overline{C}}} [L(H)] \leq& L(H,S(\overline{C})) + 2 \max_{C \in \Pi(\overline{C})}   \, \mathcal{R}(A_C,S(\overline{C})) \\
        &+ 4 \sqrt{\frac{2 \log (4/\delta)}{|S(\overline{C})|}} 
        \end{align*}    
    \end{theorem}
    
    Note that the Rademacher complexity term is at most the maximum of the Rademacher complexities of the learners used along the path from $X$ to $\overline{C}$.
    
    \newtheorem{corollary}{Corollary}[theorem]

    \begin{table*}[t!]
        \caption{Performance Comparisons for the MNIST OCR-49 Dataset, UCI Bank Marketing Dataset, and UCI Online News Popularity Dataset}    
        \label{table:all_data}
        \centering
        \begin{tabular}{ | c | c | c | c | | c | c | c | c  || c | c | c  |}
            \toprule
            \textbf{Datasets} & \multicolumn{3}{|c|}{\bf MNIST OCR-49} & \multicolumn{4}{||c||}{\bf UCI Bank Marketing}  & \multicolumn{3}{|c|}{\bf UCI Online News Popularity}   \\ \midrule
            {\bf Algorithms} & {\bf Loss}  & {\bf Gain } & {\bf p-value}  & {\bf Algorithms} &{\bf Loss}  & {\bf Gain } & {\bf p-value} & {\bf Loss}  & {\bf Gain } & {\bf p-value}  \\ \hline
            {ToPs/$\mathcal B$} & {\bf 0.0152} & -  & - &{ ToPs/$\mathcal B$} & {\bf 0.0428} &  - & - & {\bf 0.2689} & -  & -  \\ 
            {ToPs/LR} & {0.0167} & 9.0\%  & 0.187 &{ToPs/LR}  & { 0.0488} &   12.3\% & 0.089  & { 0.2801} & 4.0\% & 0.015  \\ \midrule
            Malerba (2004) \cite{ref16} & 0.0212 & 28.3\% & 0.001 &Malerba (2004) \cite{ref16} & 0.0608 & 29.5\% & $<0.001$ & 0.2994 & 10.2\% & $<0.001$ \\
            Potts (2005) \cite{ref15}& 0.0195  & 22.1\% & 0.004  &Potts (2005) \cite{ref15}& 0.0581 & 26.3\% & 0.003 & 0.2897 & 7.2\% & 0.001 \\ \midrule
            DB/Stump & 0.0177 & 14.1\% & 0.113 &AdaBoost &  0.0660 & 35.2\%& $<0.001$ & 0.3147 & 14.6\% &$<0.001$   \\
            DB/Trees$_E$  & 0.0182 & 16.5\%   & 0.114 & DTree& 0.0785 &  45.5\% & $<0.001$ & 0.3240 &  17.0\% & $<0.001$  \\
            DB/Trees$_L$ & 0.0201 &  24.4\% &0.007&DeepBoost  &  0.0591 &  27.6\% & 0.002 & 0.2929 & 8.2\% & $<0.001$ \\
            \hline
            AB/Stump1 & 0.0414 &  63.3\% & $<0.001$ &LASSO& 0.0671 &  36.2\%& $<0.001$ & 0.3209 &  16.2\% & $<0.001$ \\
            AB/Stump2 & 0.0209 & 27.3\% &0.009 &LR & 0.0669 &  36.0\%& $<0.001$ & 0.3048 & 11.8\% & $<0.001$ \\    
            AB-L1/Stump& 0.0200 & 24.0\% &0.008 &Logit &  0.0666 &  35.7\%& $<0.001$ & 0.3082 & 12.8\% & $<0.001$ \\        
            AB/Trees & 0.0198 & 23.2\% & 0.022 &LogitBoost & 0.0673 &  36.4\%& $<0.001$ & 0.3172 & 15.2\% & $<0.001$  \\
            AB-L1/Trees & 0.0197 &22.8\% &0.026 &NeuralNets & 0.0601 &  28.8\%& 0.002 & 0.2899 & 7.2\% & $<0.001$  \\
            \hline
            LB/Trees & 0.0211  & 28.0\%  &0.002 &Random Forest &  0.0548 &  21.9\% & 0.015 & 0.3074  & 12.5\%  & $<0.001$ \\
            LB-L1/Trees & 0.0201 & 24.4\% &0.009 &rbf SVM & 0.0671 &  36.2\%& $<0.001$ & 0.3081 & 12.7\% & $<0.001$   \\
             &  &  & & XGBoost & 0.0575 &  25.6\% & 0.003 & 0.3023 & 11.0\% & $<0.001$    \\
            \bottomrule
            \multicolumn{10}{l}{DB/Trees$_E$: DeepBoost with trees and exponential loss, DB/Trees$_L$: DeepBoost with trees and logistic loss. }\\
            \multicolumn{10}{l}{AB-L1/Stump: AdaBoost with stump and L1 norm. We use the same benchmarks for two UCI datasets. Bold: The best performance.}
        \end{tabular}
    \end{table*}

    \begin{corollary}
        Let $H$ be the overall predictor.  For each $\delta > 0$, with probability at least $1-\delta$,
        \begin{align*}
        \mathbb{E}_{\mathcal{D}}[L(H)]  &\leq  \frac{1}{n}\sum_{\bar{C} \in \overline{\tree}} |S(\bar{C})| \times \\
        & \Bigg[  L(H,S(\overline{C})) + 2 \max_{C \in \Pi(\overline{C})}   \, \mathcal{R}(A_C,S(\overline{C}))\\
        &+ 4 \sqrt{\frac{2 \log (4|\bar{\mathcal{T}}|/\delta)}{|S(\overline{C})|}} \Bigg]
        \end{align*}
        where $n=\sum_{\bar{C} \in \overline{\tree}} |S(\bar{C})|$. 
    \end{corollary}

    \section{Experiments}\label{sect:experiments}
    
    In this Section, we compare the performance of two instantiations of ToPs -- ToPs/LR (built on Linear Regression as the single base learner) and ToPs/${\mathcal B}$ (built on the set ${\mathcal B} =$ \{AdaBoost, Linear Regression, Logistic Regression, LogitBoost, Random Forest\} of base learners) -- against the performance of  state-of-the-art algorithms on four publicly available datasets; MNIST, Bank Marketing and Popularity of Online News datasets from UCI, and a publicly available medical dataset (survival while wait listed for a transplant). Considering two instantiations of ToPs allows us to explore the source of the improvement yielded by our method over other algorithms.  In the following subsections, we describe the datasets and the performance comparisons; the exploration of the source of improvement is illustrated in Section \ref{sect:discussion}. 
    
    We conducted 10 independent experiments with different combinations of training and testing sets; we report the mean and the standard deviations of the performances in these 10 independent experiments. To evaluate statistical significance (p-values), we assume that the prediction performances of these 10  experiments are sampled from  Gaussian distributions and we use two sample student t-tests to compute the p-value for the improvement of ToPs/${\mathcal B}$ over each benchmark.  
    
    \subsection{MNIST}\label{subsect:MNIST} 
    Here we use the MNIST OCR-49 dataset \cite{ref6}. The entire MNIST dataset consists of 70,000 samples with 400 continuous features which represent the image of a hand-written number from 0 to 9. Among 70,000 samples, we only use the 13,782 samples which represent 4 and 9; we treat 4 as label 0 (42.4\%) and 9 as label 1 (57.6\%). Each sample records all 400 features of a hand-written number image and the label of the image. There is no missing information. The objective is to classify, from the hand-written image features, whether the image represents 4 or 9. For ToPs we further divided the training samples into a training set $S$ and validation sets $V^1, V^2$ in the proportions 75\%-15\%-10\% (Same with all datasets). For comparisons, we use the results given in \cite{ref7} for various instantiations of DeepBoost, AdaBoost and LogitBoost and two model trees \cite{ref15,ref16} as   benchmarks.  To be consistent with \cite{ref7}, we use the error rate as the loss function. Table \ref{table:all_data} presents the performance comparisons: the column {\bf Gain} shows the performance gain of ToPs/$\mathcal B$ over all the other algorithms; the column {\bf p-value} shows the p-value of the statistical test (two sample student t-test) of the gain (improvement) of ToPs/${\mathcal B}$ over the other algorithms (The table is exactly as in \cite{ref7} with  ToPs/$\mathcal B$ and ToPs/LR added). ToPs/${\mathcal B}$ and ToPs/LR have 14.1\% and 5.6\% gains from the best benchmark (DeepBoost with Stump).  These improvements are not statistically significant (p-value $\sim$ 0.1) but the improvements over other machine learning methods are all statistically significant (p-values $<$ 0.05), and most are highly statistically significant (p-values $<$ 0.001).\footnote{In a separate experiment, we also compared the performance of ToPs/$\mathcal B$ on the {\em entire} MNIST dataset with that of a depth 6 CNN (AlexNet with 3 convolution nets and 3 max pooling nets). The loss of ToPs is 0.0081; this is slightly worse than CNN (0.0068) but much better than the best ensemble learning benchmark, XgBoost (0.0103). Keep in mind that CNN’s are designed for spatial tasks such as character recognition, while ToPs is a general-purpose method; as we shall see in later subsections, ToPs outperforms neural networks for other tasks.}

    \begin{table*}[tb]
        \caption{Comparison with state-of-the-art machine learning techniques. (UNOS Heart Transplant Dataset to predict wait-list mortality for different time horizons (3-month, 1-year, 3-year, and 10-year))}        
        \label{table:heart}
        \centering
        \begin{tabular}{ | c | c | c  | c | c | c| c |c | c  | c | c | c| c |}
            \toprule
            & \multicolumn{3}{|c|}{\bf 3-month mortality} & \multicolumn{3}{|c|}{\bf 1-year mortality} & \multicolumn{3}{|c|}{\bf 3-year mortality} & \multicolumn{3}{|c|}{\bf 10-year mortality}\\ \midrule
            {\bf Algorithms} & {\bf Loss} &  {\bf Gain } & {\bf p-value } &  {\bf Loss} &  {\bf Gain }& {\bf p-value }  & {\bf Loss} &  {\bf Gain } & {\bf p-value } &  {\bf Loss} &  {\bf Gain }& {\bf p-value }  \\ \midrule
            {ToPs/$\mathcal{B}$} & {\bf 0.207 } &- &- & {\bf 0.181 } &  -&  - & {\bf 0.177 } &  - &- & {\bf 0.175 } &  - &- \\
            {ToPs/LR} & {0.231 } &  10.4\% &0.006 & {0.207 } &  12.6\%& 0.014 & {0.201 } & 11.9\%& 0.004&  {0.203 } &  13.8\% &0.002\\ \midrule
            AdaBoost & 0.262 &  21.0\% & $<0.001$  & 0.239  & 24.3\% & 0.002 & 0.229 & 22.7\% & 0.001& 0.237  & 26.2\% & $<0.001$ \\
            DTree & 0.326 &  36.5\% & $<0.001$ &  0.279  & 35.1\% & $<0.001$ &  0.287  & 38.3\% & $<0.001$  & 0.249  & 29.7\% & $<0.001$  \\
            DeepBoost & 0.259 & 20.1\% & $<0.001$ & 0.245  & 26.1\% & $<0.001$& 0.219  & 19.2\% & $<0.001$  & 0.213   & 17.8\% & $<0.001$ \\
            LASSO & 0.310 &  33.2\%  & $<0.001$ & 0.281  & 35.6\% & $<0.001$ &  0.248  & 28.6\% & $<0.001$  & 0.228  & 23.2\% & $<0.001$  \\
            LR & 0.320 & 35.3\% & $<0.001$ & 0.293   & 38.2\% & $<0.001$ &  0.264  & 33.0\% & $<0.001$  & 0.285  & 38.6\% & $<0.001$ \\    
            Logit & 0.310  & 33.2\% & $<0.001$   & 0.281   & 35.6\% & $<0.001$&  0.249   & 28.9\% & $<0.001$  & 0.236  & 25.8\% & $<0.001$  \\
            LogitBoost & 0.267  & 22.5\% & $<0.001$  & 0.233  & 22.3\% & 0.002 & 0.221  & 19.9\% & 0.006 & 0.229   & 23.6\% & $<0.001$ \\
            NeuralNets & 0.262  & 21.0\% & $<0.001$  & 0.257   & 29.6\% & $<0.001$& 0.225 & 21.3\% & $<0.001$  & 0.221  & 20.8\% & $<0.001$  \\
            Random Forest & 0.252  & 17.9\% & 0.012 & 0.234  & 22.6\% & $<0.001$ & 0.217   & 18.4\% & 0.002  & 0.225  & 22.2\% & $<0.001$  \\
            SVM & 0.281   & 26.3\% & $<0.001$  & 0.243  & 25.5\% & $<0.001$ & 0.218   & 18.8\% & 0.001  & 0.284   & 38.4\% & $<0.001$ \\
            XGBoost & 0.257 &  19.5\% & $<0.001$  & 0.233  & 22.3\% & $<0.001$ & 0.223   & 20.6\% & $<0.001$ & 0.226   & 22.6\% & 0.001  \\ 
            \bottomrule        
            \multicolumn{13}{l}{Bold: The best performance.}
        \end{tabular}
    \end{table*}

    \subsection{Bank Marketing}\label{subsect:bank}       
    
    For this comparison, we use the UCI Bank Marketing dataset \cite{ref8}.  This dataset consists of 41,188 samples with 62 features; 10 of these features are continuous, and 52 are binary.  Each sample records all 62 features of a particular client and whether the client accepted a bank marketing offer (a particular term deposit account).   There is no missing information.  In this case, the objective is to predict, from the client features, whether or not the client would accept the offer.  (In the dataset 11.3\% of the clients accepted and the remaining 88.7\% declined.) To evaluate performance, we conduct 10 iterations of  5-fold cross-validation.  Because the dataset is unbalanced, we use $1 - {\rm AUC}$ as the loss function.      
    
\begin{figure*}[t!]
        \centering
        \includegraphics[width=0.65\textwidth]{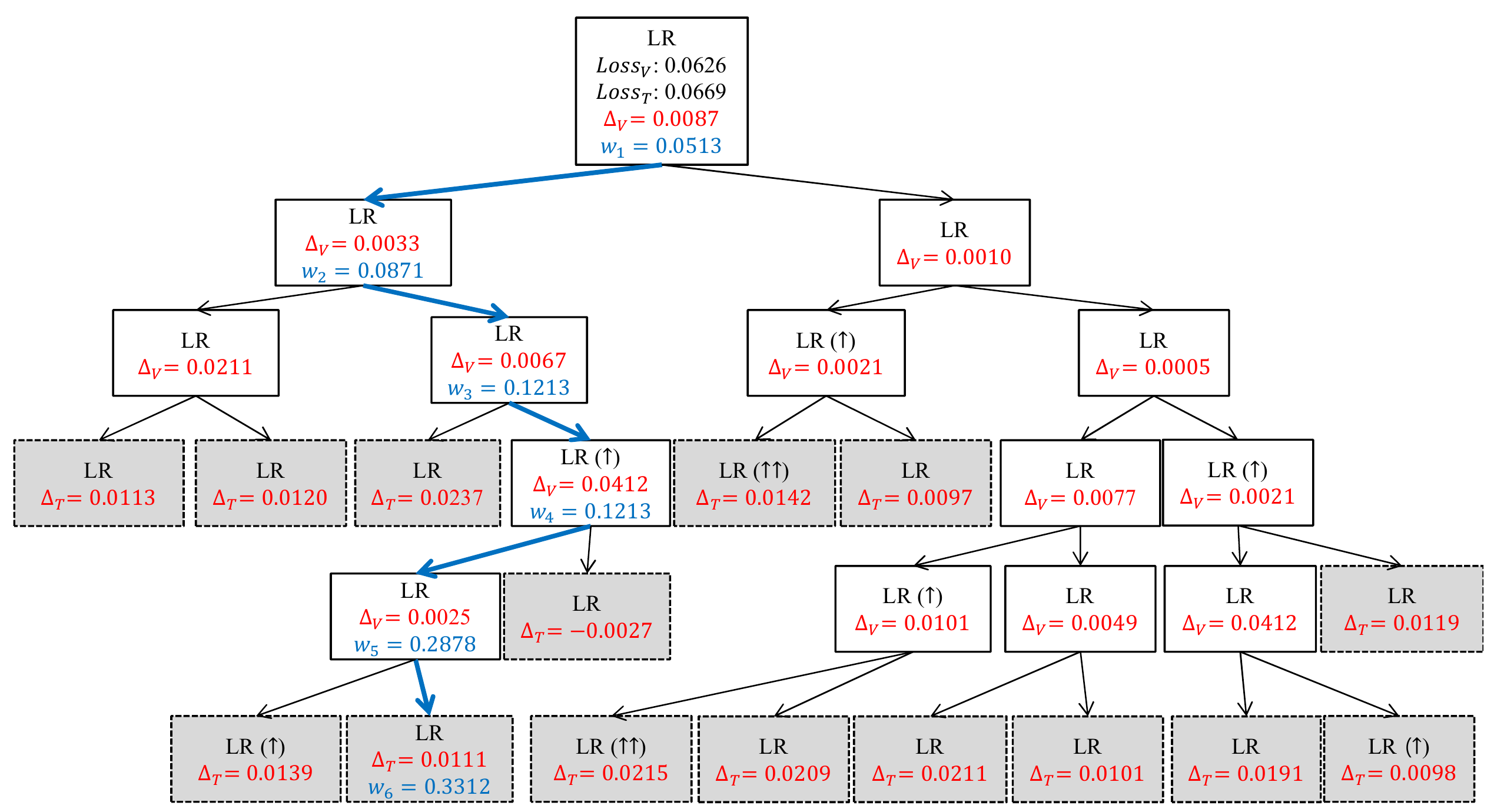}
        \caption{The locally optimal tree of predictors for ToPs/LR applied to the Bank Marketing dataset}
        \label{figure:tree_lr}
    \end{figure*}

    Table \ref{table:all_data} shows the overall performance of the our two instantiations of ToPs and  13 comparison machine-learning algorithms: Model trees \cite{ref15,ref16}, AdaBoost \cite{ref2}, Decision Trees (DT), Deep Boost \cite{ref7},  LASSO, Linear Regression (LR), Logistic Regression (Logit), LogitBoost \cite{ref9}, Neural Networks \cite{ref10}, Random Forest (RF) \cite{ref1}, Support Vector Machines (SVM) and XGBoost \cite{ref11}.  Tops/LR outperforms all of the benchmarks by more than 10\% and ToPs/$\mathcal B$  outperforms all  of the benchmarks by more than 20\%. (The improvement over Random Forest is statistically significant (p-value $=$ 0.015); the improvements over other methods are highly significant (p-values $<$ 0.005).  In Section \ref{sect:discussion} we show the locally optimal trees of predictors for this setting (Fig. \ref{figure:tree_lr} and \ref{figure:tree}) and  discuss the source of  performance gain of ToPs.

    \subsection{Popularity of Online News }\label{subsect:news}  
    Here we use the UCI Online News Popularity dataset \cite{ref8}.  This dataset consists of 39,397 samples with 57 features (43 continuous, 14  binary).  Each sample records all 57 features of a particular news item and  the number of times the item was shared.  There is no missing information. The objective is to predict, from the news features, whether or not the item would be popular -- defined to be "shared more than 5,000 times."  (In the dataset 12.8\% of the items are popular; the remaining 87.2\% are not popular.) To evaluate  performance of all algorithms, we conduct 10 iterations of  5-fold cross-validation. As it can be seen in Table \ref{table:all_data}, ToPs/LR and Tops/$\mathcal B$ both outperform all the other machine learning algorithms: Tops/$\mathcal B$ achieves a loss (measured as $1-AUC$) of 0.2689, which is 7.2\% lower than the loss achieved by the best competing benchmark. All the improvements are highly statistically significant (p-values $\leq$ 0.001).

    \subsection{Heart Transplants } 
    The UNOS (United Network for Organ Transplantation) dataset (available at  \url{https://www.unos.org/data/}) provides information about the entire cohort of 36,329 patients (in the U.S.) who were on a waiting list to receive a heart transplant but did not receive one, during the period 1985-2015.  Patients in the dataset are described by a total of 334 clinical features, but much of the feature information is missing, so we discarded 301 features for which more than 10\% of the information was missing, leaving us with 33  features -- 14 continuous and 19 binary.  To deal with the missing information, we used 10 multiple imputations using Multiple Imputation by Chained Equations (MICE) \cite{Sup15}.  In this setting, the objective(s) are to predict survival for time horizons of 3 months, 1 year, 3 years and 10 years.  (In the dataset, 68.3\% survived for 3 months, 49.5\% survived for 1 year, 28.3\% survived for 3 years, and 6.9\% survived for 10 years.) To evaluate performance, we divided the patient data based on the admission year: we took patients admitted to a waiting list in 1985-1999 as the training sample and patients admitted in 2000-2015 as the testing sample. We compared the performance of ToPs with the same machine-learning algorithms as before; the results are shown in Table \ref{table:heart}.  Once again, both ToPs/LR and ToPs/$\mathcal B$ outperform all the other machine learning algorithms: ToPs/$\mathcal B$ achieves losses (measured as $1-AUC$) between $0.175$ and $0.207$; these improve by 17.9\% to 22.3\% over the {\em best} machine learning benchmarks. The improvement over Random Forest at the 3-month horizon is statistically significant (p-value = 0.012); all other improvements are highly significant (p-values $\leq$ 0.006). The Supplementary Materials  show the  locally optimal trees of predictors for this dataset.

    \begin{figure*}[t!]
        \centering
        \includegraphics[width=0.65\textwidth]{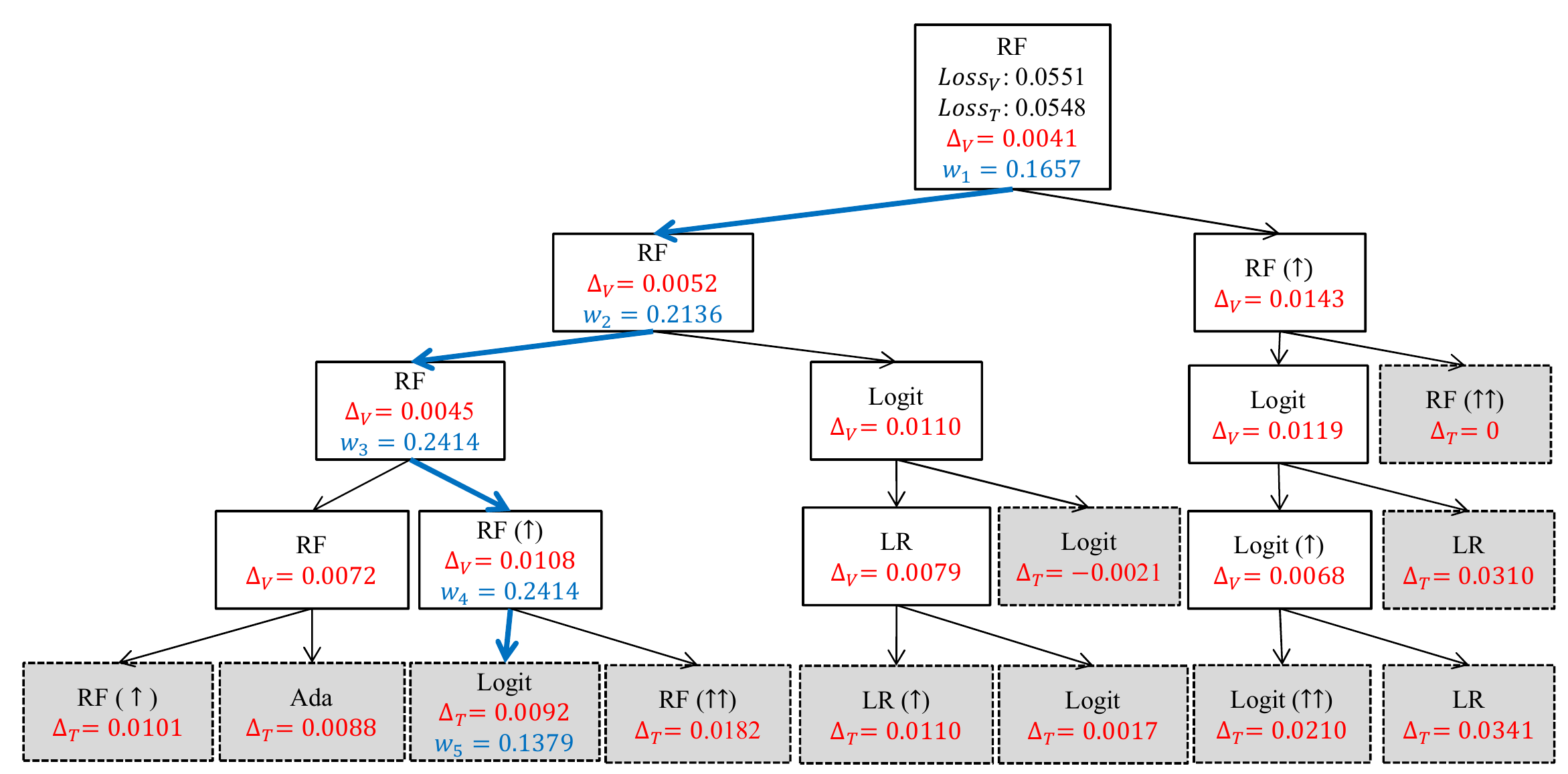}
        \caption{The locally optimal tree of predictors for ToPs/$\mathcal B$ applied to the Bank Marketing dataset}
        \label{figure:tree}
    \end{figure*}

    \section{Discussion}\label{sect:discussion}
    As the experiments show, our method yields significant performance gains over a large variety of existing machine learning algorithms.  These performance gains come from the concatenation of a number of different factors.  To aid in the discussion, we focus on the Bank Dataset and refer to Fig. \ref{figure:tree_lr} and  \ref{figure:tree}, which show the locally optimal trees of predictors grown by  ToPs/LR and ToPs/$\mathcal B$, respectively.   For ToPs/$\mathcal B$, which uses multiple learners, we show, in each node, the learner assigned to that node; for both  ToPs/LR and ToPs/$\mathcal B$ we indicate the training set  assigned to that node: For instance, in Fig. \ref{figure:tree}, Random Forest is the  learner assigned to the initial node; the training set is necessarily the entire feature space $X$.  In subsequent nodes, the training set is either the given node (if there is no marker) or the immediately preceding node (if the node is marked with a single up arrow) or the node preceding that (if the node is marked with two up arrows), and so forth.  Note that different base learners and/or different training sets are used in various nodes.  In each non-terminal node we also show the loss improvement $\Delta_v$ (computed with respect to $V^1$) obtained by splitting this node into the two immediate successor nodes.  (By construction, we split exactly when improvement is possible so     $\Delta_v $ is necessarily strictly positive at every non-terminal node while $\Delta_v $ would be $0$ at terminal nodes.)   At the terminal nodes (shaded), we show the  loss improvement $\Delta_t$ achieved on that node (that set of features) that is obtained by using the final predictor rather than the initial predictor.  Finally, for one particular path through the tree (indicated by heavy blue arrows), we show the weights assigned to the nodes along that path in computing the overall predictor. Note that the deeper nodes do not necessarily get greater weight: using the second validation set to optimize the weights compensates for overfitting deeper in the tree.

    The first key feature of our construction is that it identifies a family of subsets of the feature space -- the nodes of the locally optimal tree -- and optimally matches {\em training sets} to the nodes.  Moreover, as Fig. \ref{figure:tree_lr} and \ref{figure:tree} make clear, the optimal training set at the node $C$ need not be the node $C$ itself, but might be one of its predecessor nodes.  For ToPs/LR it is this matching of training sets to nodes that gives our method its power: if we were to use the entire training set $S$ at every node, our final predictor would reduce to simple Linear Regression -- but, as Table \ref{table:all_data} shows, because we {\em do} match training sets to nodes, the performance of ToPs/LR is 27.1\% better than that of Linear Regression.   
    
    The second key feature of our construction is that it also optimally matches {\em learners} to the nodes.  For ToPs/LR, there is only a single base learner, so the matching is trivial (LR is matched to every node) -- but for ToPs/$\mathcal B$ there are five base learners and the matching is not trivial.   Indeed, as can be seen in Fig. \ref{figure:tree}, of the five available base learners, four are actually used in the locally optimal tree.  This explains why ToPs/$\mathcal B$ improves on  ToPs/LR.  Of course, the performance of ToPs/$\mathcal B$ might be even further improved by enlarging the set of base learners.  (It seems clear that the performance of ToPs/$\mathcal A$ depends to some extent on the set of base learners $\mathcal A$.)
    
    Our recursive construction leads, in every stage, to a (potential) increase in the complexity of the predictors that can be used.  For example, Linear Regression fits a linear function to the data; ToPs/LR fits a piecewise linear function to the data.  This additional complexity raises the problem of overfitting.  However, because we train on the training set $S$ and evaluate on the first validation set $V^1$, we avoid overfitting to the training set.  As can be seen from Fig. \ref{figure:tree_lr} and \ref{figure:tree} this avoidance of overfitting is reflected in the way it limits the depth of the locally optimal tree: the growth of the tree stops when splitting no longer yields improvement {\em on the validation set} $V^1$.  Although training on $S$ and evaluating on $V^1$ avoids overfitting to $S$, it leaves open the possibility of overfitting to $V^1$; we avoid this problem by using the second validation set $V^2$ to construct optimal weights to aggregate predictions along paths. 
    
    The contrast with model trees (see again the discussion in Subsection \ref{subsect:model_trees}) is particular worth noting.  Even ToPs/LR performs significantly better than the best model trees because we assign predictive models to every node, we construct these models by training on either the current node or some parent node, we allow for splitting when only one side of the split improves performance, and we construct the final prediction as the weighted average of predictions along the path.  ToPs/$\mathcal B$ performs even better than ToPs/LR because we also allow for a richer family of base learners.

    \section{Conclusion}\label{sect:conclusion}  In this paper, we develop a new approach to ensemble learning.  We construct a locally optimal tree of predictors that matches learners and training sets to particular subsets of the feature space and aggregates these individual predictors according to endogenously determined weights.  Experiments on a variety of datasets show that this approach yields statistically significant improvements over state-of-the-art methods.

    
    %
    \appendices

    \begin{table*}[t!]
        \caption{Comparison of optimization equations with existing ensemble methods}    
        \label{table:related_work}
        \centering
        \begin{tabular}{|M|N|}
        \toprule
            \textbf{Methods} & \textbf{Optimization} \\ \midrule
            \textbf{ToPs} &  $$\min_{ \{ \mathcal{X}_1,...,\mathcal{X}_k \} } \sum_{i=1}^k \min_{h_i \in \mathcal{H} = \cup \mathcal{H}_l} \Big[ \frac{1}{N} \sum_{n=1}^N \mathcal{L}(h_i(x_j),y_j) \Big] $$ ToPs uses both multiple training sets ($\{\mathcal{X}_1,\mathcal{X}_2,...,\mathcal{X}_k\}$) and multiple hypothesis classes ($\mathcal{H}=\cup \mathcal{H}_l$) to construct the predictive models ($\{h_1,h_2,...,h_k\}$). Both multiple training sets and corresponding predictive models are jointly optimized to minimize the loss function $\mathcal{L}$. $x_j$ is the $j$-th feature, $y_j$ is the $j$-th label, $N$ is the total number of samples, and $k$  is the number of leaves.\\ \midrule
            \textbf{Bagging} &  $$\min_{h_i \in \mathcal{H}} \Big[ \frac{1}{|S|} \sum_{(x_j,y_j) \in S} \mathcal{L}(h_i(x_j),y_j) \Big] $$ Here $S$ is the family of multiple training sets, randomly drawn. Bagging uses multiple training sets (randomly drawn) and a single hypothesis class to construct multiple predictive models. Only the predictive models are optimized to minimize the loss function $(\mathcal{L})$.\\ \midrule
            \textbf{Boosting} & $$\min_{h_i \in \mathcal{H}} \Big[ \frac{1}{|S|} \sum_{(x_j,y_j) \in S} \phi_j \mathcal{L}(h_i(x_j),y_j) \Big] $$ Here $S$ is the family of multiple training sets, randomly drawn. Boosting uses multiple training sets (weighted by prediction error ($\phi_j$) of the previous model) and a single hypothesis class to construct the predictive models. Only the predictive models are optimized to minimize the loss function ($\mathcal{L}$). \\ \midrule
            \textbf{Stacking} & $$\min_{h_i \in \mathcal{H} = \cup \mathcal{H}_l} \Big[ \frac{1}{|S|} \sum_{(x_j,y_j) \in S} \mathcal{L}(h_i(x_j),y_j) \Big] $$ Stacking uses a single training set with multiple hypothesis classes to construct the predictive models. Only the predictive models are optimized to minimize the loss function ($\mathcal{L}$) . \\
            \bottomrule
        \end{tabular}
    \end{table*}

    \section{Proof of the theorem 1 (Bounds for the predictive model of each terminal node)}
    \begin{proof}
        The definition of $\mathcal{R}(\mathcal{A},S)$ is
        $$\mathcal{R}(\mathcal{A},S)=\frac{1}{|S|} \mathbb{E}_{\sigma=\{ \pm 1 \}^{|S|}} [\sup_{h \in \mathcal{A}} \sum_{z \in S} \sigma_i \times l(h,z_i) ]$$ 
        where $P(\sigma_i = 1) = P(\sigma_i = -1) = 0.5$. \\
        Let us define $\{ {A}_1,...,{A}_p\}$ as the set of all hypothesis classes that ToPs uses. \\
        Then, let us define hypothesis class $\mathcal{A}$ as follow.
        $$\mathcal{A}=\{ h=\sum_{i=1}^p w_i \times h_i | {\bf w}\in \mbox{simplex}, h_i \in A_i \}$$
        Then, 
        \begin{align*}
        \mathcal{R}(\mathcal{A},S)&=\frac{1}{|S|} \mathbb{E}_{\sigma=\{ \pm 1 \}^{|S|}} [\sup_{h \in \mathcal{A}} \sum_{z \in S} \sigma_i \times l(h,z)] \\
        &=\frac{1}{|S|} \mathbb{E}_{\sigma=\{ \pm 1 \}^{|S|}} [\sup_{h \in \mathcal{A}} \sum_{z \in S} \sigma_i \times l(\sum_{j=1}^p w_j  h_j ,z)] \\
        &=\frac{1}{|S|} \mathbb{E}_{\sigma=\{ \pm 1 \}^{|S|}} [\sup_{h \in \mathcal{A}} \sum_{z \in S} \sigma_i \times l(\sum_{j=1}^p w_j  h_j ,\sum_{j=1}^p w_j  z)] \\
        &=\frac{1}{|S|} \mathbb{E}_{\sigma=\{ \pm 1 \}^{|S|}} [\sup_{h \in \mathcal{A}} \sum_{z \in S} \sigma_i \times \hat{l}(\sum_{j=1}^p w_j  (h_j - z))] \\
        &\leq \frac{1}{|S|} \mathbb{E}_{\sigma=\{ \pm 1 \}^{|S|}} [\sup_{h \in \mathcal{A}} \sum_{z \in S} \sigma_i \times  \sum_{j=1}^p w_j  \hat{l}( h_j - z)] \\
        &= \frac{1}{|S|} \mathbb{E}_{\sigma=\{ \pm 1 \}^{|S|}} [\sup_{h \in \mathcal{A}} \sum_{z \in S} \sigma_i \times \sum_{j=1}^p w_j  {l}( h_j, z)] \\
        &\leq\frac{1}{|S|} \mathbb{E}_{\sigma=\{ \pm 1 \}^{|S|}} [\sup_{h \in \mathcal{A}} \sum_{z \in S}  \sigma_i \times \max_{j} l( h_j , z)] \\
        &= \max_{j}  \mathcal{R}(\mathcal{A}_{j},S)             
        \end{align*}
        
        Therefore, using the Rademacher Complexity Theorem,
        \begin{align*}
        \mathbb{E}_{\mathcal{D}}[L(h)]  &\leq L(h,S) + 2 \max_{i} \mathcal{R}(A_{i},S) + 4 \sqrt{\frac{2 \log (4/\delta)}{|S|}} 
        \end{align*}
        For each terminal node $\bar{C}$, the hypothesis class is $H = \sum_{C \in \Pi(\bar{C})} w^*(\Pi(\bar{C}),C) \times h_C$. Therefore, the upper bound of the expected loss for each terminal node $\bar{C}$ is
        \begin{align*}
        \mathbb{E}_{\mathcal{D}_{\bar{C}}} [L(H)] \leq &L(H,S(\bar{C})) + 2 \max_{C \in \Pi(\bar{C})}  \mathcal{R}(A_{C},S(\bar{C})) \\
        & + 4 \sqrt{\frac{2 \log (4/\delta)}{m}} 
        \end{align*}
        where $m = |S(\bar{C})|$.
    \end{proof}

    \section{Proof of the corollary 1.1 (Bounds for the entire predictive model)}
    \begin{proof}
        Based on the assumption,
        $$ L(h,Z) = \frac{1}{m} \sum_{z \in Z} l(h,z) $$ 
        where $m = |Z|$.
        Then
        $$ \mathbb{E}_{\mathcal{D}}(L(H)) = \sum_{\bar{C} \in \bar{T}} P(\bar{C}) \times \mathbb{E}_{\mathcal{D}_{\bar{C}}}[  L(H) ] $$
        Based on the Theorem 1, with probability at least $(1-\delta)^{1/|\bar{\mathcal{T}}|} $, each terminal node $\bar{C}$ satisfied the following condition.  
        \begin{align*}
        \mathbb{E}_{\mathcal{D}_{\bar{C}}}[L(H)]  \leq& L(H,S(\bar{C})) + 2 \max_{C \in \Pi(\bar{C})}  \mathcal{R}(A_{C},S(\bar{C})) \\
        &+ 4 \sqrt{\frac{2 \log (4/(1-(1-\delta)^{1/|\bar{\mathcal{T}}|}))}{m}} 
        \end{align*}
        where $m = |S(\bar{C})|$.\\
        Because the definition of the $L(h,Z)$ is the sample mean of each loss $l(h,z)$, the entire loss is the weighted average of each terminal node. (weight is the number of samples $S(|\bar{C}|)$ in each terminal node) .\\
        Furthermore, if each terminal node satisfies the above condition with probability at least $(1-\delta)^{1/|\bar{\mathcal{T}}|}$, it means that the probability that all terminal nodes satisfies the following condition is at least $ ((1-\delta)^{1/|\bar{\mathcal{T}}|})^{|\bar{\mathcal{T}}|} = (1-\delta)$\\
        Therefore, with at least $1-\delta$ probability,
        \begin{align*}
        \mathbb{E}_{\mathcal{D}}[L(H)]  \leq& \sum_{\bar{C} \in \bar{T}} \frac{|S(\bar{C})|}{n} L(H,S(\bar{C})) \\
        &+ 2 \sum_{\bar{C} \in \bar{T}} \frac{|S(\bar{C})|}{n} \max_{C \in \Pi(\bar{C})}   \mathcal{R}(A_{C},S(\bar{C})) \\
        &+ 4 \sum_{\bar{C} \in \bar{T}} \frac{|S(\bar{C})|}{n}  \sqrt{\frac{2 \log[4/(1-(1-\delta)^{1/|\bar{\mathcal{T}}|})]}{|S(\bar{C})|}} 
        \end{align*}
        Furthermore, $1-(1-\delta)^{1/|\bar{\mathcal{T}}|} \geq \frac{\delta}{|\bar{\mathcal{T}}|}$. Thus, we can switch $1-(1-\delta)^{1/|\bar{\mathcal{T}}|}$ to $\frac{\delta}{|\bar{\mathcal{T}}|}$ in this inequality. (Using Binomial Series Theorem, the inequality is easily proved.) Therefore, with at least $1-\delta$ probability,
        \begin{align*}
        \mathbb{E}_{\mathcal{D}}[L(H)]  \leq&  \frac{1}{n}\sum_{\bar{C} \in \overline{\tree}} |S(\bar{C})| \times \\
        &\Bigg[  L(H,S(\overline{C})) + 2 \max_{C \in \Pi(\overline{C})}   \, \mathcal{R}(A_C,S(\overline{C}))\\
        &+ 4 \sqrt{\frac{2 \log (4|\bar{\mathcal{T}}|/\delta)}{|S(\overline{C})|}} \Bigg]
        \end{align*}
    \end{proof}
    
    \section{Proof of the computational complexity}
    
    \subsection{Proof of computational complexity of one recursive step: }
    
    \textbf{Statement: The computational complexity of one recursive step for constructing tree of predictors grows as $\sum_{i=1}^M \mathcal{O}(ND \times T_i(N,D) )$.}
    
    \begin{proof}
        There are two procedures in one recursive steps of constructing a tree of predictors. (1) Greedy search for the division point, (2) Construct the predictive model for each division. 
        
        First, the possible combinations of dividing the feature space into two subspaces using one feature with the threshold are at most $N \times D$ where $N$ is the total number of samples, and $D$ is the number of dimensions. Because, in each subspace, there should be at least one sample.
        
        Second, for each division, we need to construct $M$ predictive models. The computational complexity to construct the $M$ number of predictive models for each division is trivially computed as $\sum_{i=1}^M \mathcal{O}(T_i(N,D))$ where $\mathcal{O}(T_i(N,D))$ is the computational complexity of $i$-th learner to construct the predictive model with $N$ samples and $D$ dimensional features.
        
        Therefore, the computational complexity of one recursive step with $M$ learner can be written as follow. 
        $$ND \times \sum_{i=1}^M \mathcal{O}(T_i(N,D)) = \sum_{i=1}^M \mathcal{O}(ND \times T_i(N,D))$$
        Because, there are $ND$ possibilities of division and for each division, the computational complexity is $\sum_{i=1}^M \mathcal{O}(T_i(N,D))$.
        
    \end{proof}
    
    \subsection{Proof of computational complexity of constructing the entire tree of predictors}
    
    \textbf{Statement: The computational complexity of constructing the entire tree of predictors is $\sum_{i=1}^M \mathcal{O}(N^2 D \times T_i(N,D))$}
    
    \begin{proof}
        By the previous statement, we know that the computational complexity of one recursive step of constructing the tree of predictors is $\sum_{i=1}^M \mathcal{O}(ND \times T_i(N,D))$. Therefore, now, we need to figure out the maximum number of recursive steps with $N$ samples. 
        
        For each recursive step, the number of clusters is stepwise increased. Furthermore, in each cluster, there should be at least one sample. Therefore, we can easily figure out that the maximum number of recursive steps are at most $N$. (Note that the recursive steps are much less needed in practice because ToPs does not divide all the samples into different clusters) 
        
        Therefore, the computational complexity of the entire construction of tree of predictors is as follow.
        $$N \times \sum_{i=1}^M \mathcal{O}(ND \times T_i(N,D)) = \sum_{i=1}^M \mathcal{O}(N^2 D \times T_i(N,D))$$
    \end{proof}
    For instance, if we only use the linear regression as the learner of ToPs (ToPs/LR), the computational complexity of ToPs is $\mathcal{O}(N^3 D^3)$. This is because, the computational complexity of linear regression is $\mathcal{O}(N D^2)$. Therefore, the entire computational complexity of ToPs/LR is $\sum_{i=1}^M \mathcal{O}(N^{2} D T_i(n,D)) = \mathcal{O}(N^3 D^3)$

    \section*{Acknowledgment}
    This work was supported by the Office of Naval Research (ONR) and the NSF (Grant number: ECCS1462245).

    \ifCLASSOPTIONcaptionsoff
    \newpage
    \fi

    
    
    %
    %


\end{document}